\documentclass[10pt,twocolumn,letterpaper]{article}

\usepackage{cvpr}              
\usepackage{multirow}

\definecolor{cvprblue}{rgb}{0.21,0.49,0.74}
\usepackage[pagebackref,breaklinks,colorlinks,allcolors=cvprblue]{hyperref}

\def\paperID{1294} 
\def\confName{CVPR}
\def\confYear{2025}

\title{Hierarchical Multi-Graphs Learning for Robust Group Re-Identification}

\author{
Ruiqi Liu\textsuperscript{1}, Xingyu Liu\textsuperscript{2}, Xiaohao Xu\textsuperscript{3}, Yixuan Zhang\textsuperscript{1}, Yongxin Ge\textsuperscript{2}, Lubin Weng\textsuperscript{1}\\
\textsuperscript{1}College of Artificial Intelligence, University of Chinese Academy of Sciences, Beijing, China\\
\textsuperscript{2}College of Big Data and Software, Chongqing University, Chongqing, China\\
\textsuperscript{3}University of Michigan - Ann Arbor\\
}

\begin{document}
\maketitle
\begin{abstract}
Group Re-identification (G-ReID) faces greater complexity than individual Re-identification (ReID), due to challenges like mutual occlusion, dynamic member interactions, and evolving group structures. Prior graph-based approaches have aimed to capture these dynamics by modeling the group as a single topological structure. However, these methods struggle to generalize across diverse group compositions, as they fail to fully represent the multifaceted relationships within the group.
In this study, we introduce a \textit{Hierarchical Multi-Graphs Learning }(\textbf{HMGL}) framework to address these challenges. Our approach models the group as a collection of multi-relational graphs, leveraging both explicit features—such as occlusion, appearance, and foreground information—and implicit dependencies between members. This hierarchical representation, which is encoded via  \textit{Multi-Graphs Neural  Network }(\textbf{MGNN}), allows us to resolve ambiguities in member relationships, particularly in complex, densely populated scenes. To further enhance matching accuracy, we propose a \textit{Multi-Scale Matching} (\textbf{MSM}) algorithm, which mitigates issues of member information ambiguity and sensitivity to hard samples, improving robustness in challenging scenarios.
Our method achieves state-of-the-art performance on two standard benchmarks (CSG and RoadGroup), with Rank-1/mAP scores of 95.3\%/94.4\% and 93.9\%/95.4\%, respectively, marking a notable  1.7\% and 2.5\% improvement in Rank-1 accuracy over existing approaches.

\end{abstract}
    
\section{Introduction}

In recent years, Re-Identification (ReID) in video surveillance has attracted considerable attention \cite{wang2014region,wang2017statistical,wu2022learning,respondek2019counting,bai2017scalable}, particularly as many social activities occur in groups, making group monitoring essential. {Group Re-Identification (G-ReID)} is a critical task that involves identifying and matching groups, typically consisting of 2 to 6 individuals, across multiple camera views. Groups are considered identical if they share at least 60\% of their members. However, G-ReID poses more complex challenges than individual ReID due to dynamic membership changes, varying group layout, and occlusions that obscure individual features and complicate intra-group relationships \cite{cai2010matching,lin2019group,lisanti2017group,ristani2016performance,zheng2009associating}. These challenges are compounded by the subtle, often hidden connections within groups—such as appearance similarities, shared motion patterns, and behavioral cues—that are crucial for linking individuals across different views, yet difficult to capture with traditional methods.


\begin{figure}[t!]
	\centering
	\includegraphics[scale=0.07]{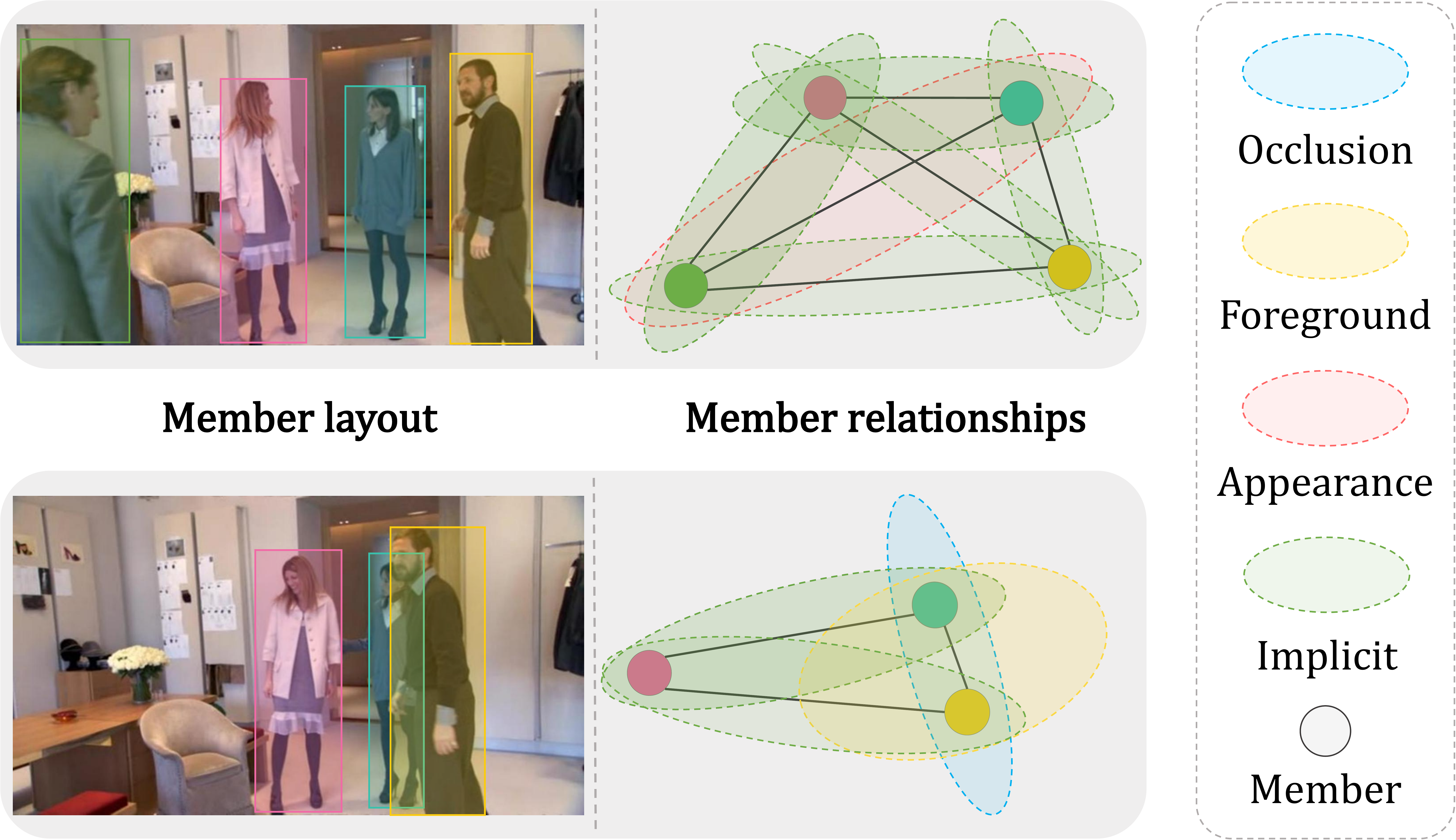}
	\caption{\textbf{Motivation illustration}. Groups exhibit intricate relationships among members, including occlusion, foreground, appearance, and less obvious implicit connections. These densely intertwined relationships challenge the ability of a single graph to comprehensively represent the group structure, necessitating a more robust, multi-graphs approach.}
	\label{fig_1} 
\end{figure}


Existing graph-based G-ReID approaches model groups as combinations of individuals and their relationships, using graph structures to address dynamic layouts and changing memberships \cite{huang2019dot,zhu2020group,yan2020learning}. Although these methods have made progress, several key issues remain unresolved. Firstly, most existing approaches rely on a single graph structure to represent the group, which is insufficient to capture the complexity of group dynamics. As illustrated in Fig. \ref{fig_1}, mutual occlusions and shifting group arrangements lead to diverse and intricate member relationships, limiting the representational power of a single graph. Furthermore, occlusions obscure vital visual cues, making it challenging to maintain consistent identity information across different scenarios. While some work has attempted to mitigate the effects of occlusion \cite{lu2022tsn,lu2023siamese,yu2023patch}, these methods often isolate the occluded members without fully recognizing the novel relationships that arise from such interactions.

Another underexplored area in G-ReID is group matching during the inference phase. Existing methods can be broadly classified into two categories: {group-level matching} and {member-level matching}. Group-level matching aggregates the information of individual members to form a representation of the entire group, but this often blurs individual details, leading to suboptimal accuracy. member-level matching, on the other hand, searches for the best combination of matching members between groups, yielding higher accuracy but remaining susceptible to interference from difficult samples.


To address these challenges, we propose a {Hierarchical Multi-Graphs Learning  (HMGL)} framework that models groups as combinations of members and their various relationships. Our framework constructs {multi-relational awareness graphs} that preserve relationship-specific attributes during node information propagation. For instance, nodes with occlusion relationships to an anchor node can help reduce interference in the anchor node’s features, while appearance-based connections can provide rich contextual information. We use a learnable affinity matrix to model the group as a global awareness graph, which is decoupled into explicit relationship graphs (occlusion, foreground, appearance) and an implicit relationship graph. A {Multi-Graphs Neural Network (MGNN)} is then employed to learn robust group member features, with multi-graphs convolution layers that generate embeddings integrating diverse relational contexts, leading to more accurate feature extraction.


Complementing our framework, we introduce a {Multi-Scale Matching (MSM)} algorithm to overcome the limitations of existing group matching strategies. Our method constructs matching objective functions at the node, subgraph, and whole graph levels, allowing group similarity measurement on individual, local, and global scales. By framing both individual- and local-scale matching as the bipartite graph matching problems, MSM effectively addresses the complex relationships within groups, enhancing robustness against challenging samples and leading to improved retrieval performance compared to existing approaches.

The main contributions of this work are:

\begin{itemize}
    \item We introduce a novel {Hierarchical Multi-Graphs Learning  (HMGL)} framework that models groups using both explicit and implicit relationship-aware graphs. Additionally, we design a {Multi-Graphs Neural Network (MGNN)} that fully explores the multiple dependencies within the group, enabling more precise feature extraction.
    
    \item We propose an advanced {Multi-Scale Matching  (MSM)} algorithm that overcomes the limitations of traditional matching algorithms by reducing member information ambiguity and sensitivity to challenging samples through the construction of multi-scales bipartite graph matching.
    
    \item Our method achieves state-of-the-art performance on the CSG and RoadGroup datasets, achieving Rank-1/mAP scores of 95.3\%/94.4\% and 93.9\%/95.4\%, respectively.
\end{itemize}

\section{Related Work}
\subsection{Group Re-Identification}.
In the field of video surveillance and security, the task of Group Re-Identification (G-ReID) is full of challenges, but research on G-ReID is relatively scarce ~\cite{lisanti2017group,ristani2016performance,zheng2009associating}. Some pioneering works tackled this task by extracting global or semi-global features. Covariance~\cite{cai2010matching} used discriminative covariance descriptors, and CRRRO-BRO~\cite{zheng2009associating} segmented group images into annular regions to extract statistical features. However, both methods were overly sensitive to layout changes, limiting recognition performance. To fully utilize individual features, PREF~\cite{lisanti2017group} and BSC+CM~\cite{zhu2016consistent} introduced patch matching between group photos, but this requires well-vertically aligned group images.
With the rise of deep learning, methods based on Convolutional Neural Networks (CNN) and Transformers ~\cite{vaswani2017attention} have become mainstream. DotGNN~\cite{huang2019dot} adopted CycleGAN for style transfer combined with GNN for group feature extraction; GCGNN~\cite{zhu2020group} used a spatial K-nearest neighbors graph to model group relationships, designing a group context GNN; MACG~\cite{yan2020learning} proposed a multi-attention context graph for optimizing group feature learning. SOT~\cite{zhang2022uncertainty} and UMSOT~\cite{zhang2024uncertainty} combined uncertainty modeling with Transformers for better decision boundaries, while 3DT~\cite{zhang2022modeling} used Transformer models to reconstruct 3D spatial relationships. Despite technical advances, these works lack exploration of complex intra-group relationships. In this work, we model the group as a multi-relational graph to thoroughly explore internal dependencies.


\subsection{Group Matching}
Unlike ReID, G-ReID requires group-level matching, complicated by varying group sizes and member arrangements. Some works simplify matching through group-level features: DotGNN~\cite{huang2019dot} uses pooling after graph learning; MACG~\cite{yan2020learning} employs attention to aggregate representative individuals;.BGFNet~\cite{wang2022learning} enhances graph layout features via information propagation. However, these aggregation methods blur member details.
Other works use permutation-invariant models ~\cite{zaheer2017deep,lee2019set,murphy2018janossy,fei2022dumlp,wu2021representing,zhang2021deep} for member-level matching. For example, GCGNN~\cite{zhu2020group} and MHD~\cite{yu2023patch} use the Hungarian-based algorithm for member's feature matching based on cosine similarity; CPM~\cite{xiong2023similarity} calculates closest permutation distance for all query-gallery image pairs. However, these methods often fail to account for challenging sample interference in complex group matching.
LIMI~\cite{xiao2018group} delve into finer-grained member relationships by introducing  multi-order matching at various granularities within groups to calculate similarity. However, this multi-granularity modeling is limited by the quantitative relationships among members and cannot capture the characteristics of different scales within groups in matching. We adopt a similarity measure that considers complex intra-group relationships and propose a multi-scale group matching algorithm utilizing graph relationship clustering.

\begin{figure*}[t]
	\centering
	\includegraphics[scale=0.13]{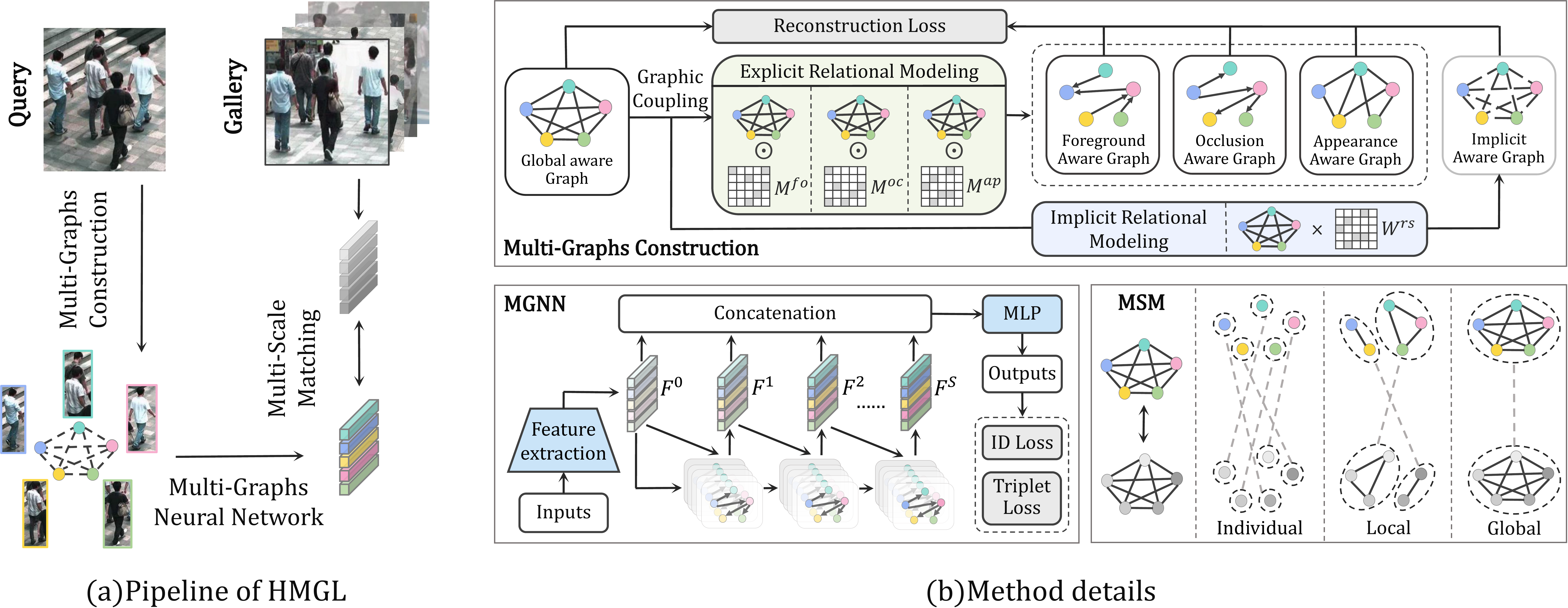}
 \caption{
\textbf{Hierarchical Multi-Graphs Learning Framework for Group Re-Identification.} The framework models groups as multi-relational graphs, incorporating both explicit (e.g., foreground, occlusion, appearance) and implicit relationships to capture complex dependencies within group images. A reconstruction loss $\mathcal{L}^{re}$ is applied to enhance the representational capacity of these relationship-aware graphs. The Multi-Graphs Neural Network(MGNN) then explores the various relational dependencies within groups, yielding robust feature representations for each group member. Finally, group similarity is computed using the Multi-Scale Matching(MSM) approach, enabling effective comparison and re-identification across group instances.
}
	\label{fig_2}
\end{figure*}
\section{Problem Formulation of G-ReID}

The goal of group re-identification (G-ReID) is to match groups of multiple agents across different camera views by learning robust group representations. Given multi-view images rather than predefined groups, this problem can be divided into two main subproblems: group construction and group matching.

\noindent\textbf{Group construction.} multi-view images are processed to identify and cluster agents into groups based on spatial, visual, or contextual cues, resulting in a set of constructed groups $\mathcal{G} = \{G_1, G_2, \ldots, G_M\}$. Each constructed group $G_i$ is assigned an identity label $y_i \in \{1, \ldots, K\}$, where $K$ denotes the total number of unique group identities in the dataset. Each group $G_i$ consists of $n_i$ agents, represented by $\{a_{i1}, a_{i2}, \ldots, a_{in_i}\}$.

\noindent\textbf{Group matching.} In this phase,  our objective is to learn a mapping function $f: G_i \rightarrow \mathcal{F}_i$ that embeds each group $G_i$ into a feature space $\mathcal{F}_i$, ensuring that groups with the same identity are mapped to proximate regions while groups with distinct identities remain separated:
\begin{equation}
d(f(G_i), f(G_j)) \leq \epsilon, \quad \text{if } y_i = y_j,
\end{equation}
\begin{equation}
d(f(G_i), f(G_j)) > \epsilon, \quad \text{if } y_i \neq y_j,
\end{equation}
where $d(\cdot, \cdot)$ denotes a distance metric (e.g., Euclidean distance), and $\epsilon$ is a threshold defining the matching proximity in the feature space.

\section{Our Method}

\noindent \textbf{Pipeline overview.} We propose a novel framework for G-ReID that captures complex intra-group relationships and enhances group matching. As illustrated in Fig. \ref{fig_2}, our method follows three key steps: (1) Multi-relational awareness graphs are constructed by extracting explicit (occlusion, appearance, foreground) and implicit information from group images. (2) A Multi-Graphs Neural Network (MGNN) propagates contextual information across these graphs, generating robust member representations. (3) A Multi-Scale Matching (MSM) algorithm conducts bipartite graph matching across multiple scales, enabling reliable inter-group similarity measurement. Detailed descriptions of each component are provided in the following sections.


\subsection{Multi-relational Awareness Graphs}

With the introduction of GNNs, graphs are commonly used to model dependencies between connected nodes, and past work has typically described group memberships with a single graph\cite{defferrard2016convolutional,duvenaud2015convolutional,ktena2017distance,niepert2016learning,qi2018learning}. However, group dynamics such as membership shifts, layout changes, and mutual occlusions create diverse and complex relationships within a crowd. To address conflicts in information dissemination within a single graph structure, we propose constructing multi-relational awareness graphs for each group.

Specifically, for a given group $I$ containing $N$ members $\left\{I_1,I_2,\ldots,I_N\right\}$, we define a global-aware graph $\mathcal{G}^0=\left\{\mathcal{V},\mathcal{E}^0\right\}$ to model the overall relationships within the group , each node $i\in\mathcal{V}$ corresponds to the appearance characteristics of the corresponding member $I_i$ in the group. The affinity matrix $\mathcal{A}^0\in\mathbb{R}^{N\times N}$ of $\mathcal{G}^0$ is formulated as:
\begin{small}
\begin{equation}
\begin{split}
\mathcal{A}^0&={D^0}^{-1}{\widetilde{\mathcal{A}}}^0 \vspace{0.3ex}, 
D_{(i,i)}^0=\sum_{j}{\widetilde{\mathcal{A}}}_{(i,j)}^0, \\
{\widetilde{\mathcal{A}}}^0&=exp\left(-\frac{{\widetilde{W}}^{0^2}}{2\sigma^2}\right), {\widetilde{W}}^0=F^0\left(\frac{W^0+{W^0}^T}{2}\right){F^0}^T,
\end{split}
\end{equation}
\end{small}
where $\sigma$ is the standard deviation of ${\widetilde{W}}^0$, $F^0\in\mathbb{R}^{N\times d}$ denotes the initial features of each node within the graph, and $W^0\in\mathbb{R}^{N\times N}$ is a learnable parameter. The multi-relational awareness graphs will be obtained based on the decoupling of the global-aware graph $\mathcal{G}^0$ as follows.

\noindent\textbf{Explicit relational modeling}. On the one hand, since propagating appearance information between individuals enhances feature robustness, we propose the construction of appearance-aware graphs $\mathcal{G}^{ap}=\left\{\mathcal{V},\mathcal{E}^{ap}\right\}$ to capture appearance-based relationships.  For any two appearance-similar nodes, an edge $e_{ij}\in\mathcal{E}^{ap}$ is retained from node $i$ to node $j$. We define an appearance-aware masking matrix $M^{ap}\in\mathbb{R}^{N\times N}$ to filter out edges unrelated to appearance.
\begin{equation}
\begin{split}
M_{\left(i,j\right)}^{ap}=\left\{\begin{matrix}1,\ \ &\text{if}\ D_{\left(i,j\right)}^{ap}>\tau
 \\0,\ \ &\ \ \text{otherwise}
\end{matrix}\right. ,\\
D^{ap}=-Norm\left(\mathcal{O}\left(F^0,F^0\right)\right),
\end{split}
\end{equation}
where $\tau$ denotes the appearance similarity threshold, $\mathcal{O}\left(\cdot,\cdot\right)$ denotes the Euclidean distance matrix of the two embedded sets, and $Norm\left(\cdot\right)$ denotes the hierarchical normalization.

On the other hand, since individuals walking in pairs often occlude each other, resulting in visual feature masking and new foreground associations, we propose constructing occlusion-aware and foreground-aware graphs, $\mathcal{G}^{oc}=\left\{\mathcal{V},\mathcal{E}^{oc}\right\}$ and $\mathcal{G}^{fo }=\left\{\mathcal{V},\mathcal{E}^{fo}\right\}$, to model mutual occlusions. For two nodes $i$ and $j$ in the graph, if node $i$ occludes node $j$, the directed edges $e_{ij}\in\mathcal{E}^{oc}$ and $e_{ji}\in\mathcal{E}^{fo}$ are preserved for node $i$ to node $j$. Specifically, we construct the occlusion-aware masking matrix $M^{oc}$ and the foreground-aware masking matrix $M^{fo}$ to represent this process:
\begin{equation}
\begin{split}
M_{\left(i,j\right)}^{oc}=M_{\left(j,i\right)}^{fo}=\left\{\begin{matrix}1, &\text{if}\ i\in occlusion\left(j\right)
 \\0,&\text{otherwise}
\end{matrix}\right.,
\end{split}
\end{equation}
where $occlusion(j)$ denotes the set of all nodes that occlude node $j$. The occlusion and foreground labels are  generated based on whether the bounding boxes of a pair of individuals overlap and the relative distribution of human keypoints. We define the affinity matrices $\mathcal{A}^{ap}$, $\mathcal{A}^{oc}$ and $\mathcal{A}^{fo}$ for $\mathcal{G}^{ap}$, $\mathcal{G}^{oc}$ and $\mathcal{G}^{fo}$ as:
\begin{equation}
    \begin{split}
        \mathcal{A}^{*}&={D^{*}}^{-1}{\widetilde{\mathcal{A}}}^{*}\odot M^{*},\\
        D_{(i,i)}^{*}&=\sum_{j}{\widetilde{\mathcal{A}}}_{(i,j)}^{*},
        {\widetilde{\mathcal{A}}}^{*}=W^{*}{\widetilde{\mathcal{A}}}^0,
    \end{split}
\end{equation}
where $^{*}$ represent $^{ap/oc/fo}$, $W^{ap}\in\mathbb{R}^{N\times N}$ is the learnable appearance-aware matrix, $W^{oc}$ and $W^{fo}$ are the learnable occlusion-aware matrix and foreground-aware matrix respectively. 


\noindent\textbf{Implicit relational modeling}. Beyond intuitively perceived membership relationships, groups often contain deeper and less obvious topological connections.  To capture these implicit dependencies and reduce information loss in the graph decoupling process, we construct an implicit-aware graph  $\mathcal{G}^{rs}$ to model hidden relationships within the group. The affinity matrix $\mathcal{A}^{rs}$ of $\mathcal{G}^{rs}$ is formulated as:
\begin{equation}
\begin{split}
    \mathcal{A}^{rs}&={D^{rs}}^{-1}{\widetilde{\mathcal{A}}}^{rs},\\
    D_{(i,i)}^{rs}&=\sum_{j}{\widetilde{\mathcal{A}}}_{(i,j)}^{rs},
    {\widetilde{\mathcal{A}}}^{rs}=W^{rs}{\widetilde{\mathcal{A}}}^0,
\end{split}
\end{equation}
where $W^{rs}$ is the implicit-aware learning matrix. The implicit-aware graph enables the propagation of subtle yet critical group relationships, reinforcing intrinsic connections that are not immediately observable. This enhances adaptability to complex environments and strengthens awareness of individual interactions and influences.

Finally, to enhance the relational information in the global-aware graph and improve multi-relational awareness graphs representation, we define the reconstruction loss $\mathcal{L}^{re}$ on the multi-relational awareness graphs to constrain the decoupling process of the global-aware graph:
\begin{equation}
    \begin{split}
        \mathcal{L}^{re}&=\sqrt{\sum_{i=1}^{m}\sum_{j=1}^{n}\left(\mathcal{A}_{ij}^0-\mathcal{A}_{ij}^\prime\right)^2},\\
        \mathcal{A}^\prime&=(\mathcal{A}^{ap}+\mathcal{A}^{oc}+\mathcal{A}^{fo}+\mathcal{A}^{rs})W^{re},
    \end{split}
\end{equation}
where $W^{re}$ is the a learnable graph transformation matrix.

\subsection{Multi-Graphs Neural Network (MGNN)}
Appearance-aware, occlusion-aware, foreground-aware, and implicit-aware graphs reflect the associations between nodes from different perspectives, and in order to integrate multi-relational information, we deploy a Multi-Graphs Neural Network(MGNN) on the multi-relational awareness graphs. Specifically, member images are processed through a pre-trained feature extractor $E$ on ImageNet \cite{2009ImageNet} to obtain the initial feature representation $F^0 \in \mathbb{R}^{N \times d}$:
\begin{equation}
F^0=E\left(I_1,I_2,\ldots,I_N\right).
\end{equation}

After extracting member feature representations, we use these as initial node features to the multi-relational awareness graphs for representation learning.  In each graph convolution layer, we concatenate the node features with the graph convolution outputs of each relational awareness graph, updating the nodes for the next layer as:
\begin{equation}
F^s=\left[F^0;\left(\mathcal{A}^0;\mathcal{A}^{ap};\mathcal{A}^{oc};\mathcal{A}^{fo};\mathcal{A}^{rs}\right)F^{s-1}\right]W^{dim}.
\end{equation}

Finally, the weighted sum of node features from each graph convolution layer is used as the feature output result $F^{out}\in\mathbb{R}^{N\times out}$, which is formulated as:
\begin{equation}
       F^{out}=\sum_{s=0}^{S}F^sW_s^{out},
\end{equation}
where $W^{out}\in\mathbb{R}^{(S+1)\times N\times out}$ and $W^{dim}\in\mathbb{R}^{6d\times d}$ are both learnable dimensional transformation matrices, and $S$ is the number of layers in the graph convolution.
 
While population matching relies on accurate node feature extraction, accurate prediction of node labels is a meaningful goal. After $S$ rounds multi-graphs convolution, we use a classifier for node label prediction and add identity loss $\mathcal{L}^{id}$ and hard sample triplet loss $\mathcal{L}^{trip}$ to constrain node features. The process is given by:
\begin{equation}
        \mathcal{L}^{id}=\sum_{i=1}^{N}{y_i^{true}log{\left(y_i^{pred}\right)}},
\end{equation}
\begin{equation}
        \mathcal{L}^{trip}=\sum_{i=1}^{N}\left\{0,m+\mathop{max}\limits_{p\in{Pos}_i}{dist_{ip}}-\mathop{min}\limits_{n\in{Neg}_i}{dist_{in}}\right\},
\end{equation}
where $y_i^{true}$ denotes the true label of node $i$, $dist$ denotes the Euclidean distance between node features, ${Pos}_i$ and ${Neg}_i$ denote the positive and negative sample sets of node $i$ in the same batch, respectively. Therefore, the total loss of our proposed  HMGL during training is
\begin{equation}
\mathcal{L}^{all}=\mathcal{L}^{id}+\mathcal{L}^{trip}+\delta\mathcal{L}^{re}.
\end{equation}

\subsection{Multi-Scale Matching (MSM)}
In group matching, our goal is to compute the similarity score between a query group image and all the  gallery images.  Given a query set ${I_{q1}, I_{q2}, \ldots, I_{qN}}$ and a gallery set ${I_{g1}, I_{g2}, \ldots, I_{gM}}$, we construct global-aware graphs $\mathcal{G}_q^0$ and $\mathcal{G}_g^0$, and use MGNN to extract node embeddings, $F_q^{out} \in \mathbb{R}^{N\times out}$ and $F_g^{out} \in \mathbb{R}^{M\times out}$.  To improve group matching, we introduce a multi-scale approach that balances individual-, local-, and global-scale matching. Both individual-scale and local-scale matching are treated as bipartite graph optimal matching problems to mitigate hard sample effects.

\noindent\textbf{Individual-scale matching}. We construct the global-aware graphs $\mathcal{G}_q^0$ and $\mathcal{G}_g^0$ as bipartite graphs $\mathcal{G}^{nod}=\left\{\mathcal{V}_q^{nod}\cup\mathcal{V}_g^{nod},\mathcal{E}^{nod} \right\}$, whose affinity matrix $\mathcal{A}^{nod}$ is formulated as:
\begin{equation}
    \begin{split}
        \mathcal{A}_{\left(i,j\right)}^{nod}&=\left\{\begin{matrix}D_{\left(i,j\right)}^{nod},&\text{if}\ i\in\mathcal{V}_q\ \text{and}\ j\in\mathcal{V}_g\\0,&\text{otherwise}\end{matrix}\right.,\\
        D^{nod}&=ReLu\left(-Norm\left(\mathcal{O}\left(F_q^{out},F_g^{out}\right)\right)\right).
    \end{split}
    \label{Individual-scale}
\end{equation}

Subsequently, we solve for the best matching edge weights and $\mathcal{P}^{nod}$ as a measure of the similarity scores of node-level matches on the bipartite graph $\mathcal{G}^{nod}$:
\begin{equation}
        \mathcal{P}^{nod}=KM\left(\mathcal{G}^{nod}\right),
\end{equation}
where $KM\left(\cdot\right)$ denotes bipartite graph best matching.

\begin{table*}[t]
    \centering
     
   \caption{Performance comparison of the proposed method with state-of-the-art techniques on the CSG and RoadGroup datasets. } 
    \resizebox{\textwidth}{!}{
    \begin{tabular}{l|ccccc|ccccc}
\hline
\multirow{2}{*}{Method}  & \multicolumn{5}{c|}{CSG} & \multicolumn{5}{c}{RoadGroup} \\ \cline{2-11} 
                        &Rank-1&Rank-5&Rank-10&Rank-20&mAP&Rank-1&Rank-5&Rank-10&Rank-20&mAP       \\ \hline
          CRRRO-BRO~\cite{zheng2009associating}      &10.4 &25.8 &37.5&51.2&-          &17.8   &34.6   &48.1  &62.2  &-      \\
         Covariance~\cite{cai2010matching}     &16.5 &34.1 &47.9&67.0&-          &38.0   &61.0   &73.1  &82.5  &-      \\
            PREF~\cite{lisanti2017group}        &19.2 &36.4 &51.8&70.7&-          &43.0   &68.7   &77.9  &85.2  &-      \\
            BSC+CM~\cite{zhu2016consistent}       &24.6 &38.5 &55.1&73.8&-          &58.6   &80.6   &87.4  &92.1  &-      \\ \hline
            DoTGNN~\cite{huang2019dot}      &-    &-    &-   &-   &-          &74.1   &90.1   &92.6  &98.8  &-      \\ 
            GCGNN~\cite{zhu2020group}        &-    &-    &-   &-   &-          &81.7   &94.3   &96.5  &97.8  &-      \\
            MACG~\cite{yan2020learning}        &63.2 &75.4 &79.7&84.4&-          &84.5   &95.1   &96.9  &98.1  &-      \\ \hline
            LIMI~\cite{xiao2018group}  &-    &-    &-   &-   &-          &72.3   &90.6   &94.1  &-     &-      \\
            DotSCN~\cite{huang2020dotscn}  &-    &-    &-   &-   &-          &84.0   &95.1   &96.3  &-     &-      \\
            MGR~\cite{lin2019group}         &57.8 &71.6 &76.5&82.3&-          &80.2   &93.8   &96.3  &97.5  &-      \\
            3DT~\cite{zhang2022modeling}         &92.9 &97.3 &98.1&-  &92.1           &91.4   &97.5   &98.8  &-  &94.3      \\
            SOT~\cite{zhang2022uncertainty}         &91.7 &96.5 &97.6&-  &90.7           &86.4   &96.3   &98.8  &-  &91.3      \\
            UMSOT~\cite{zhang2024uncertainty}       &93.6 &97.3 &98.3&-  &92.6           &88.9   &95.1   &98.8  &-  &91.7      \\\hline

            \textbf{HMGL (Ours)}         &\textbf{95.3} &\textbf{97.7} &\textbf{98.3}&\textbf{98.9}&\textbf{94.4}       &\textbf{93.9}   &\textbf{98.4}   &\textbf{99.0}  &\textbf{99.2}  &\textbf{95.4}      \\ \hline

\end{tabular}}
    \label{compare}
\end{table*}

\noindent\textbf{Local-scale matching}. Benefiting from the propagation of contextual information in multi-graphs learning, the global-aware graphs $\mathcal{G}_q^0$ and $\mathcal{G}_g^0$ have acquired comprehensive relational information. We propose to perform spectral clustering on their affinity matrices $\mathcal{A}_q^0$ and $\mathcal{A}_g^0$ for local-scale matching respectively. Specifically, the normalized symmetric Laplace matrix is firstly derived:
\begin{equation}
    \begin{split}
        L_{q/g}&=I-D_{q/g}^{-{1/2}}\mathcal{A}_{q/g}^0D_{q/g}^{-{1/2}}, \\D_{(i,i)}&=\sum_{j}\mathcal{A}_{\left(i,j\right)}^0.
    \end{split}
\end{equation}

After that, the eigenvector matrix $U_{q/g}\in\mathbb{R}^{N/M\times k}$ corresponding to the $t\left (1<t<min\left(N,M\right)\right)$ minimal non-zero eigenvalues of $L_{q/g}$ is computed. We clustered them after row normalization to obtain the subsets $\mathcal{V}_q^{sub}$ and $\mathcal{V}_g^{sub}$:
\begin{equation}
    \begin{split}
        \mathcal{V}_{q/g}^{sub}&=\overline{KMeans}\left(U_{q/g}^\prime\right),
        U_i^\prime=\frac{U_i}{\Vert U_i \Vert}, 
    \end{split}
\end{equation}
where $\overline{KMeans}\left(\cdot\right)$ denotes that only the centroids from the K-Means clustering results are retained. Subsequently, as defined in Eq. (\ref{Individual-scale}), we construct the subgraph-level bipartite graph $\mathcal{G}^{sub}=\left\{\mathcal{V}_q^{sub}\cup\mathcal{V}_g^{sub}, \mathcal{E}^{sub}\right\}$ and solve for the optimal matching edge weights, yielding $\mathcal{P}^{sub}=KM\left(\mathcal{G}^{sub}\right)$, which serves as the similarity score for local-scale matching. In cases where $min\left(N,M\right)\le2$, the local-scale matching process is skipped.

\noindent\textbf{Global-scale matching}. The global-level matching similarity score $\mathcal{S}^{glo}$ between $\mathcal{G}_q^0$ and $\mathcal{G}_g^0$ is:
\begin{equation}
        \mathcal{P}^{glo}=\Vert {\bar{F}}_q^{out}-{\bar{F}}_g^{out} \Vert,
\end{equation}
where ${\bar{F}}_q^{out}$ and ${\bar{F}}_g^{out}$ are the feature centers of the node embedding sets $F_q^{out}$ and $F_g^{out}$.

Specifically, the multi-scale objective function for group matching can be characterized as:
\begin{equation}
        \mathcal{P}=\alpha\mathcal{P}^{nod}+\beta\mathcal{P}^{sub}+\gamma\mathcal{P}^{glo},
\end{equation}
where $\mathcal{P}^{nod}, \mathcal{P}^{sub}$, $\mathcal{P}^{glo}$ are the individual-scale matching target, local-scale matching target and global-scale matching target, respectively. $\alpha$, $\beta$, and $\gamma$ are hyperparameters controlling the weights at different scales.
\section{Experiments}





\subsection{Dataset and Evaluation Metrics}

We evaluated our method on two public G-ReID datasets:

\noindent\textbf{RoadGroup} ~\cite{lin2019group} consists of images captured by two non-overlapping cameras positioned along a busy road. This dataset includes 162 group identities, with each group represented by two cropped images.

\noindent\textbf{CUHK-SYSU Group (CSG)} ~\cite{yan2020learning} contains 3839 group images across 1558 group identities and approximately 3500 member identities. The images are sourced from both surveillance footage and non-surveillance contexts, such as movies and TV shows.

\noindent\textbf{Dataset preprocessing.} In accordance with the protocol outlined in \cite{lin2019group}, we randomly and equally divided the group images from the RoadGroup dataset into training and testing sets. For the CSG dataset, following the approach in \cite{yan2020learning}, we used 859 groups for training and the remaining 699 groups for testing. To prepare the images of individual members, we cropped each member from the group images based on the provided bounding boxes and resized them to 256 $\times$ 128 pixels.

\noindent\textbf{Metrics}. we use Cumulative Matching Characteristics (CMC) and mean Average Precision (mAP).

\subsection{Implementation Details}

Our model utilizes ResNet50 ~\cite{he2016deep}, pre-trained on ImageNet ~\cite{2009ImageNet}, as the backbone for the member feature transformer. We set the weight of reconstruction loss $\delta=0.2$ , the number of multi-graphs convolution layers $S=2$, the number of clusters for local-scale matching $t=3$, and assign the multi-scale matching weighting parameters $\alpha=0.6$, $\beta=0.3$, and $\gamma=0.1$. Training is performed with the SGD optimizer ~\cite{bottou2012stochastic}, using an initial learning rate of 0.0003, a maximum of 200 epochs, and a batch size of 16. All ablation studies, parameter analyses, and visualizations were conducted on the CSG dataset.

\subsection{Comparison with the State-of-the-Arts}

We evaluated the performance of our method against several state-of-the-art approaches across two public datasets, organizing them into two main categories: hand-crafted feature-based methods and deep learning methods. The hand-crafted methods included CRRRO-BRO ~\cite{zheng2009associating}, Covariance ~\cite{cai2010matching}, PREF ~\cite{lisanti2017group}, and BSC+CM ~\cite{zhu2016consistent}. Among deep learning methods, we further classified them by the use of graph structures: graph-based methods comprised DOTGNN ~\cite{huang2019dot}, GCGNN ~\cite{zhu2020group}, and MACG ~\cite{yan2020learning}, while non-graph-based methods included LIMI ~\cite{xiao2018group}, MGR~\cite{lin2019group}, and SOT ~\cite{zhang2022uncertainty}.

\noindent\textbf{Quantitative results.}
As shown in Table \ref{compare}, our proposed  HMGL, with ResNet50 as the backbone, achieved superior performance on both the CSG and RoadGroup datasets, with Rank-1/mAP of 95.3\%/94.4\% and 93.9\%/95.4\%, respectively. On the CSG dataset,  HMGL surpassed the best-performing UMSOT method by 1.7\% in Rank-1 accuracy. Similarly, on the RoadGroup dataset, HMGL outperformed the top-performing 3DT by 2.5\% in Rank-1 accuracy. Additionally,  HMGL achieved a substantial improvement over all previous graph-based methods, indicating that the use of multi-relational graphs enhances group modeling by overcoming the limitations of single-graph structures, thereby significantly increasing representational capacity.

\begin{figure}[t!]
	\centering  
	\includegraphics[scale=0.082]{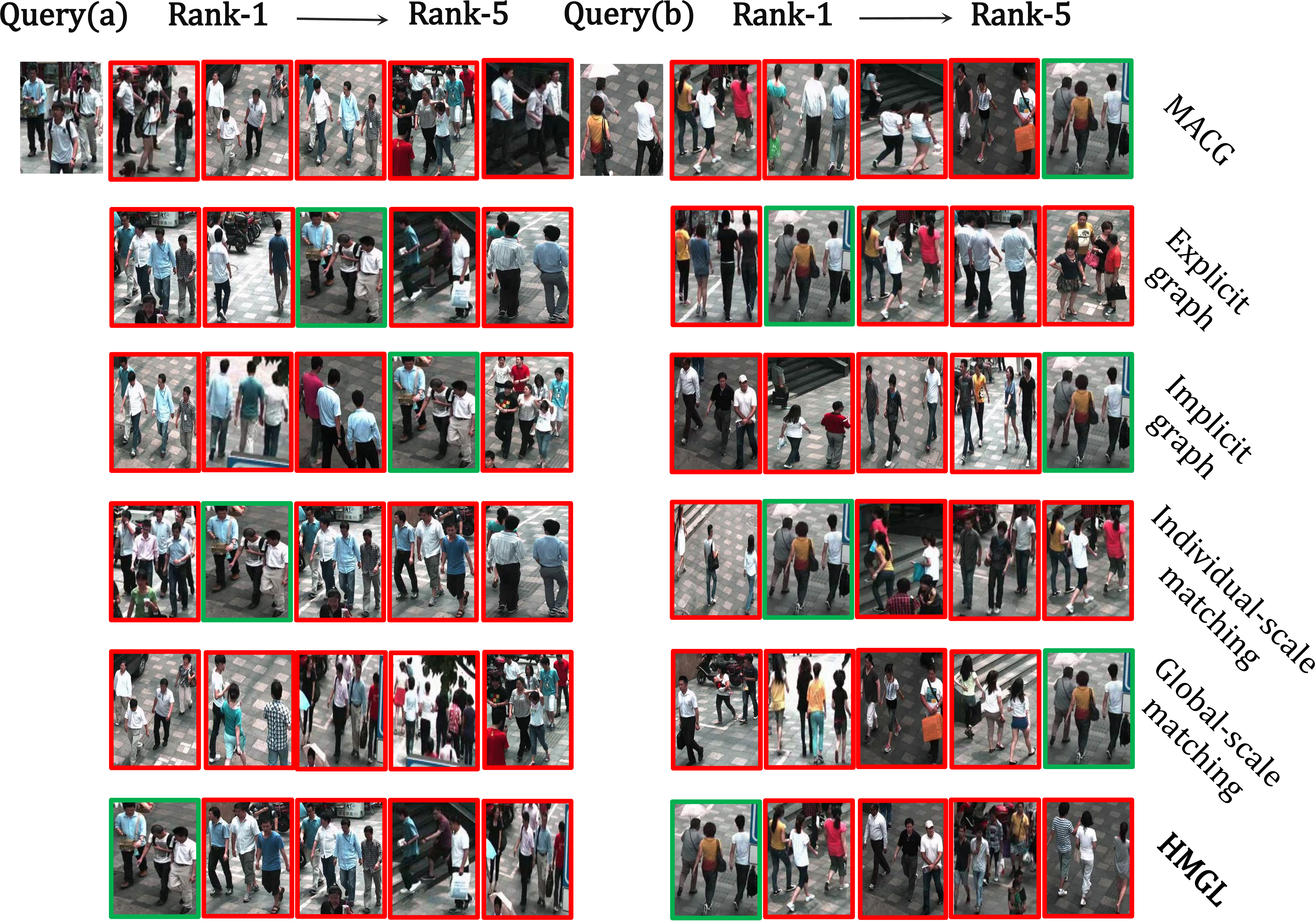}
	\caption{ \textbf{Comparative analysis of group re-identification methods based on Rank-1 to Rank-5 retrieval performance.} Two query examples (a and b) are shown on the left for each method, with retrieved images ranked in descending order of similarity. Green-bordered images indicate correctly matched groups, while red-bordered images indicate incorrect matches. } 
	\label{fig_3}
   
\end{figure}

\begin{table}[t]
    \centering  
    \caption{Ablation study on each relational awareness graph.}
    \begin{tabular}{c|c|c|c|c|c|c}
\hline
\multirow{2}{*}{NO.} & \multicolumn{4}{c|}{Components} & \multicolumn{2}{c}{CSG Dataset}      \\ \cline{2-7} 
               &$\mathcal{G}^{ap}$ & $\mathcal{G}^{oc}$ & $\mathcal{G}^{fo}$ & $\mathcal{G}^{rs}$ & Rank-1 & mAP \\  \hline \hline
    1          & $\times$      & $\times$  & $\times$  & $\times$  &85.6 &84.6     \\
    2          & $\checkmark$  & $\times$  & $\times$  & $\times$  &83.1 &82.3     \\
    3          & $\times$  & $\checkmark$  & $\times$  & $\times$  &88.6 &87.3     \\ 
    4          & $\times$  & $\times$  & $\checkmark$  & $\times$  &87.7 &87.1     \\ 
    5          & $\times$  & $\times$  & $\times$  & $\checkmark$  &90.7 &89.9     \\
    6          & $\times$  & $\checkmark$  & $\checkmark$  & $\times$  &91.2 &90.6     \\ 
    7          & $\checkmark$  & $\checkmark$  & $\checkmark$  & $\times$  &92.3 &91.5     \\ \hline
    8          & $\checkmark$  & $\checkmark$  & $\checkmark$  & $\checkmark$ &\textbf{95.3} &\textbf{94.4}  \\ \hline
\end{tabular}
    \label{tab_1} \vspace{-5mm}
\end{table}

\noindent\textbf{Qualitative results.} 
As shown in Figure \ref{fig_3}, queries a and b are presented as retrieval examples to analyze the performance of methods based on different graph constructions and group matching under extreme occlusion and dynamic group changes. MACG, recognized as a classic method incorporating both graph-based and global matching techniques, did not yield satisfactory results. In the graph construction tests, the isolated use of explicit or implicit graph methods demonstrated considerable instability across diverse samples. In the group matching tests, global matching methods struggled to handle difficult samples, while node matching methods showed some effectiveness but still did not reach ideal performance. In contrast, our approach demonstrated stable and outstanding performance in matching difficult samples.


\subsection{Ablation Study}

\begin{figure}[t!]
	\centering
	\includegraphics[scale=0.105]{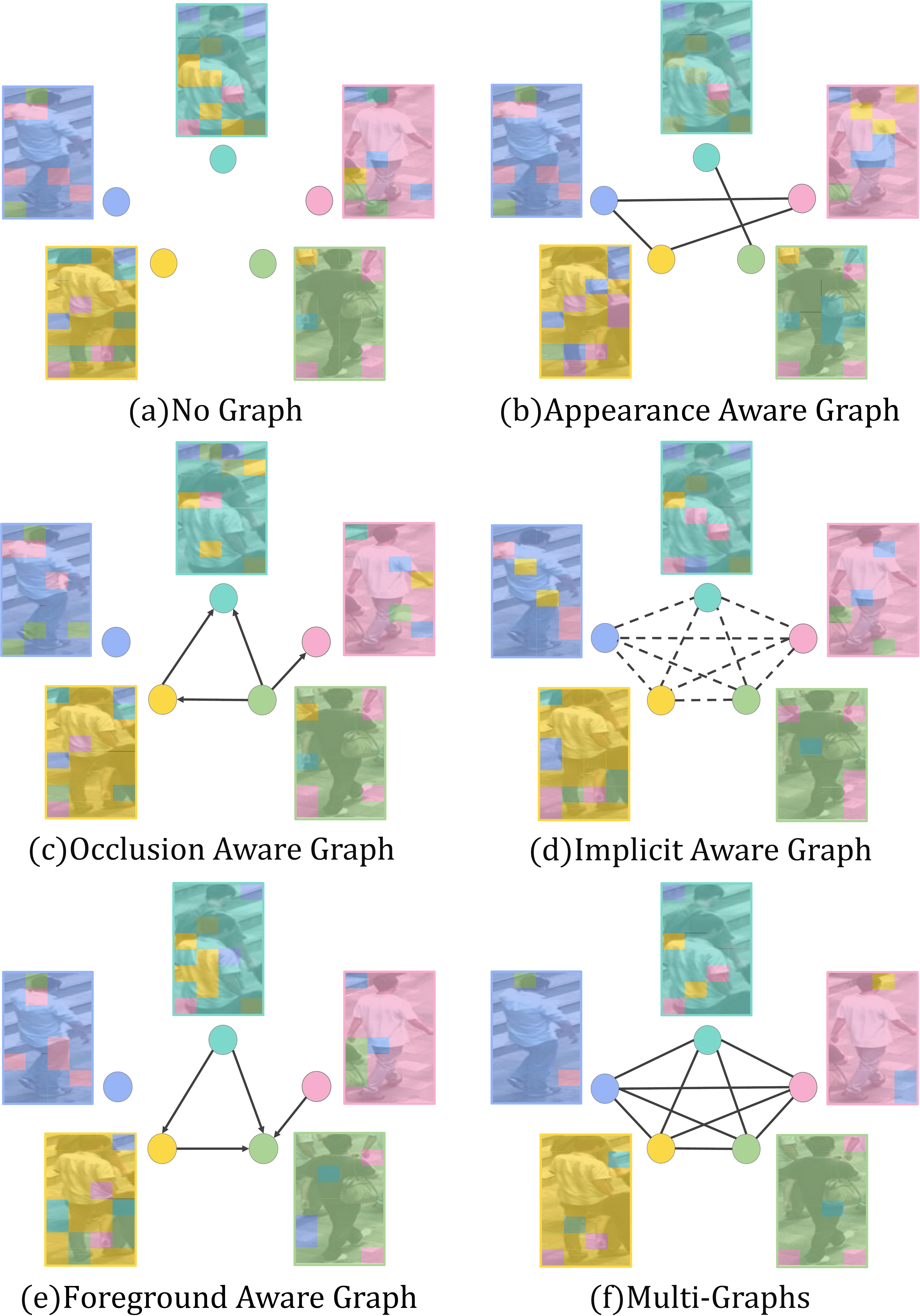}
	\caption{\textbf{Visualization of various graph-based models for group re-identification}\textbf{, showcasing the structure and connectivity in each approach,} alongside an analysis of feature map inconsistencies among group members.  Each member is represented by an 8$\times$4 feature vector from the last convolutional layer, with a global average pooling vector representing individual identities. Ideally, each section of a member’s feature map should align uniquely with their own representation, but some regions display similarity to other members’ representations, indicating potential anomalies in the group re-identification process.}
	\label{fig_4} 
\end{figure}

\noindent\textbf{Effectiveness of multi-relational awareness graphs}. To analyze the impact of different explicit and implicit relationship-aware graphs in modeling group relationships, we experimented with various combinations of the appearance, occlusion, foreground, and implicit-aware graph convolution layers in multi-graphs learning. As observed in Table \ref{tab_1}, independent occlusion-aware, foreground-aware, and implicit relational graphs each contributed to performance improvements, with occlusion-aware and implicit graphs showing particularly significant contributions. This validates the importance of occlusion relationship and implicit relationship in G-ReID. Notably, the independent use of the appearance-aware graph led to performance degradation, while its combination with other relational graphs improved performance. We speculate that the independent appearance-aware graph may cause feature ambiguity among similar members, while other relational graphs mitigate this issue and retain effective appearance information for group modeling. Figure \ref{fig_4} visualizes the impact of different relational graph combinations on the extraction of member feature maps, where we argue that an ideal feature map should exhibit stronger internal consistency. By introducing various specific relational information, multi-relational graphs model groups from multiple perspectives, helping the network learn more robust features.

\noindent\textbf{Effectiveness of loss functions}. To individually verify the necessity of identity loss $\mathcal{L}^{id}$, triplet loss $\mathcal{L}^{trip}$, and graph reconstruction loss $\mathcal{L}^{re}$, we sequentially disabled one loss function from participating in gradient computation while keeping the other two. As shown in Table \ref{tab_2}, incorporating identity loss and triplet loss ensures a reliable basic performance of the model, while adding graph reconstruction loss leads to a significant improvement in performance.  This is due to the proposed graph reconstruction loss imposing constraints on generating multiple relationship aware graphs, which helps mitigate information loss during the global graph decoupling process and enriches the expressive capability of each relationship aware graph.

\begin{table}[t]
    \centering 
    \caption{Ablation study of loss functions.}
    \begin{tabular}{c|c|c|c|c|c}
\hline
\multirow{2}{*}{NO.} & \multicolumn{3}{c|}{Components}   & \multicolumn{2}{c}{CSG Dataset}      \\ \cline{2-6} 
               & $\mathcal{L}^{id}$ & $\mathcal{L}^{trip}$ & $\mathcal{L}^{re}$ & Rank-1 & mAP \\  \hline \hline
    1          & $\times$  & $\checkmark$  & $\checkmark$  &80.9 &79.8     \\
    2          & $\checkmark$  & $\times$  & $\checkmark$  &87.5 &86.1     \\ 
    3          & $\checkmark$  & $\checkmark$  & $\times$  &92.7 &91.2     \\\hline
    4          & $\checkmark$  & $\checkmark$  & $\checkmark$ &\textbf{95.3} &\textbf{94.4}  \\ \hline
\end{tabular}
    \label{tab_2} 
\end{table}

\begin{table}[t]
    \centering  
    \caption{Ablation study of matching effects in MSM.}
    \begin{tabular}{c|c|c|c|c|c}
\hline
\multirow{2}{*}{NO.} & \multicolumn{3}{c|}{Components}   & \multicolumn{2}{c}{CSG Dataset}      \\ \cline{2-6} 
               & $\mathcal{P}^{nod}$ & $\mathcal{P}^{sub}$ & $\mathcal{P}^{glo}$  & Rank-1 & mAP \\  \hline \hline
    1          & $\times$  & $\times$  & $\checkmark$  &67.2   &65.6     \\ 
    2          & $\checkmark$  & $\times$  & $\times$  &83.3  &82.6     \\
    3          & $\checkmark$  & $\times$  & $\checkmark$  &86.7     &85.5     \\ 
    4          & $\checkmark$  & $\checkmark$  & $\times$  &93.9     &92.8     \\ \hline
    5          & $\checkmark$  & $\checkmark$  & $\checkmark$ &\textbf{95.3} &\textbf{94.4}   \\ \hline
\end{tabular}
    \label{tab_3} 
\end{table}

\noindent\textbf{Effectiveness of Multi-Scale Matching (MSM) }. To verify the contribution of individual-scale matching, local-scale matching, and global-scale matching, we sequentially removed these three modules from the group matching objective $\mathcal{P}$. As can be seen from Table \ref{tab_3}, it is difficult to obtain impressive matching results only by relying on whole-graph features, and both individual-scale matching and local-scale matching have a significant impact on the final matching results. In particular, individual-scale matching is the greatest guarantee of matching accuracy, while local-scale matching and global-scale matching overcome the impact of some difficult samples, further optimizing performance.

\subsection{Hyper-parameter Analysis}
To examine the effect of key hyper-parameters, we conducted parameter analysis experiments on the CSG dataset.

\begin{figure}[t]
	\centering
	\includegraphics[scale=0.32]{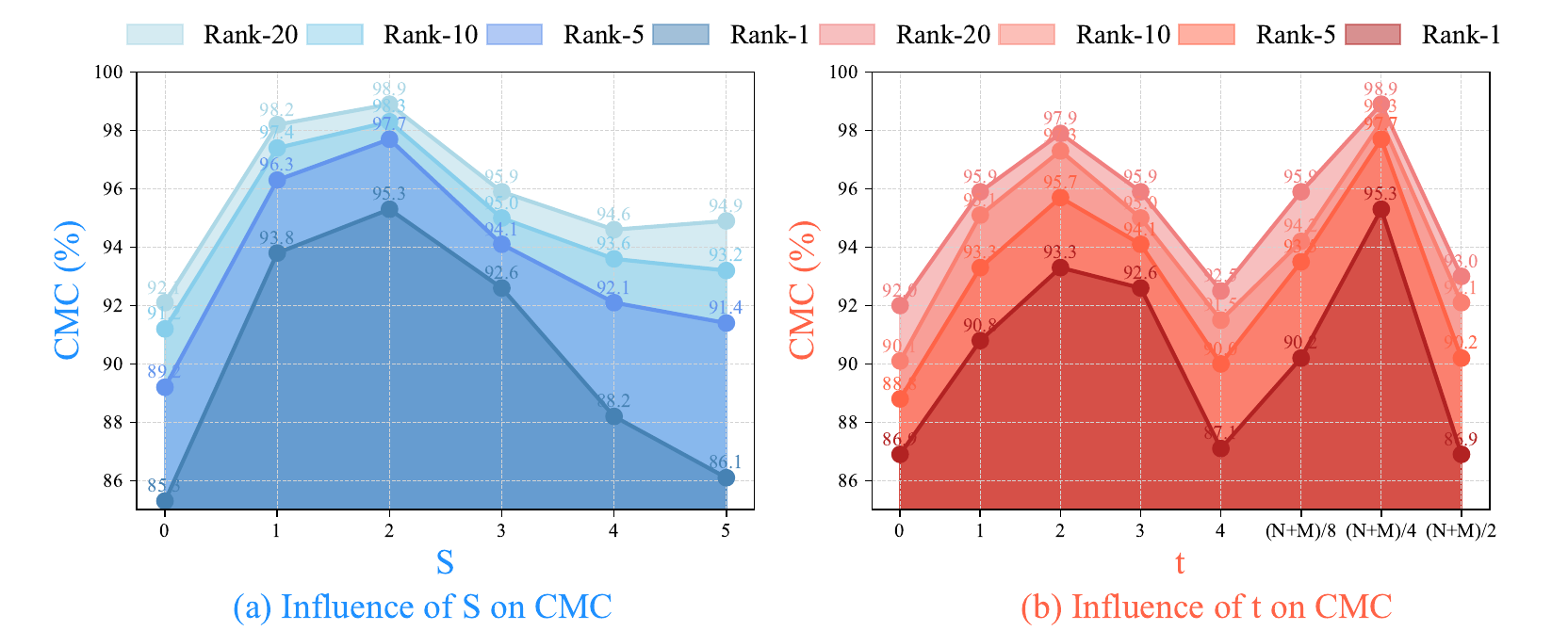}
	\caption{Analysis of the impact of key parameters on the CMC at Rank-1, Rank-5, Rank-10, and Rank-20.}
	\label{fig_5}
\end{figure}

\noindent\textbf{Effect of multi-graphs convolution layer numder $S$}. As shown in Fig.\ref{fig_5}(a). Setting $S=0$ (using backbone features directly) leaves the model vulnerable to occlusions and group changes, yielding suboptimal results. As $S$ increases, richer inter-node associations improve performance, peaking at $S=2$. However, too many layers lead to node homogenization within the group, reducing performance.

\noindent\textbf{Effect of subgraph number $t$}. The number of clusters for local-scale matching affects segmentation granularity. A small $t$ results in overly large subgraphs, resembling global-scale matching, while a large $t$ overlaps with individual-scale matching. To adapt to varying node counts $N$ and $M$ across groups, we tested fixed values and values influenced by $N$ and $M$. Fig.\ref{fig_5}(b) shows that when $t \approx \frac{N+M}{4}$, Tighter intra-subgraph node relationships and sparser inter-subgraph edges yield optimal matching performance.

\section{Conclusion}
We have introduced a Hierarchical Multi-Graphs Learning Framework (HMGL) to address the complexities of Group Re-Identification (G-ReID), effectively capturing intricate intra-group relationships through multi-relational awareness graphs and a dedicated Multi-Graphs Neural Network (MGNN). Our approach handles challenges like inter-member occlusions and dynamic interactions, while the proposed Multi-Scale Matching (MSM) Algorithm enhances group matching precision, achieving strong performance on public benchmarks.

\noindent\textbf{Future work}. HMGL faces challenges in handling complex intra-group relationships, where information transfer may be incomplete. Future work will focus on refining the multi-relational graph structure to better capture these intricate dependencies, particularly in large-scale and dynamic environments. Additionally, we aim to extend our method to broader applications, including real-world multi-agent scenarios and diverse group settings, to further validate its robustness and adaptability across diverse conditions.

{
    \small
    \bibliographystyle{ieeenat_fullname}
    \bibliography{main}

\begin{thebibliography}{40}
\providecommand{\natexlab}[1]{#1}
\providecommand{\url}[1]{\texttt{#1}}
\expandafter\ifx\csname urlstyle\endcsname\relax
  \providecommand{\doi}[1]{doi: #1}\else
  \providecommand{\doi}{doi: \begingroup \urlstyle{rm}\Url}\fi

\bibitem[Bai et~al.(2017)Bai, Bai, and Tian]{bai2017scalable}
Song Bai, Xiang Bai, and Qi Tian.
\newblock Scalable person re-identification on supervised smoothed manifold.
\newblock In \emph{CVPR}, pages 2530--2539, 2017.

\bibitem[Bottou(2012)]{bottou2012stochastic}
L{\'e}on Bottou.
\newblock Stochastic gradient descent tricks.
\newblock In \emph{Neural Networks: Tricks of the Trade: Second Edition}, pages 421--436. Springer, 2012.

\bibitem[Cai et~al.(2010)Cai, Takala, and Pietikainen]{cai2010matching}
Yinghao Cai, Valtteri Takala, and Matti Pietikainen.
\newblock Matching groups of people by covariance descriptor.
\newblock In \emph{ICPR}, pages 2744--2747. IEEE, 2010.

\bibitem[Contributors(2024)]{ultralytics2024}
Ultralytics Contributors.
\newblock Ultralytics.
\newblock \url{https://github.com/ultralytics/ultralytics}, 2024.

\bibitem[Defferrard et~al.(2016)Defferrard, Bresson, and Vandergheynst]{defferrard2016convolutional}
Micha{\"e}l Defferrard, Xavier Bresson, and Pierre Vandergheynst.
\newblock Convolutional neural networks on graphs with fast localized spectral filtering.
\newblock \emph{NeurIPS}, 29, 2016.

\bibitem[Deng et~al.(2009)Deng, Dong, Socher, Li, Li, and Fei-Fei]{2009ImageNet}
Jia Deng, Wei Dong, R. Socher, Li~Jia Li, Kai Li, and Li Fei-Fei.
\newblock Imagenet: A large-scale hierarchical image database.
\newblock \emph{Proc of IEEE Computer Vision \& Pattern Recognition}, pages 248--255, 2009.

\bibitem[Duvenaud et~al.(2015)Duvenaud, Maclaurin, Iparraguirre, Bombarell, Hirzel, Aspuru-Guzik, and Adams]{duvenaud2015convolutional}
David~K Duvenaud, Dougal Maclaurin, Jorge Iparraguirre, Rafael Bombarell, Timothy Hirzel, Al{\'a}n Aspuru-Guzik, and Ryan~P Adams.
\newblock Convolutional networks on graphs for learning molecular fingerprints.
\newblock \emph{NeurIPS}, 28, 2015.

\bibitem[Fei et~al.(2022)Fei, Zhu, Liu, Deng, Li, Deng, and Zhang]{fei2022dumlp}
Jiajun Fei, Ziyu Zhu, Wenlei Liu, Zhidong Deng, Mingyang Li, Huanjun Deng, and Shuo Zhang.
\newblock Dumlp-pin: a dual-mlp-dot-product permutation-invariant network for set feature extraction.
\newblock In \emph{AAAI}, pages 598--606, 2022.

\bibitem[He et~al.(2016)He, Zhang, Ren, and Sun]{he2016deep}
Kaiming He, Xiangyu Zhang, Shaoqing Ren, and Jian Sun.
\newblock Deep residual learning for image recognition.
\newblock In \emph{CVPR}, pages 770--778, 2016.

\bibitem[Huang et~al.(2019)Huang, Wang, Hu, Lin, and Satoh]{huang2019dot}
Ziling Huang, Zheng Wang, Wei Hu, Chia-Wen Lin, and Shin'ichi Satoh.
\newblock Dot-gnn: Domain-transferred graph neural network for group re-identification.
\newblock In \emph{ACM MM}, pages 1888--1896, 2019.

\bibitem[Huang et~al.(2020)Huang, Wang, Tsai, Satoh, and Lin]{huang2020dotscn}
Ziling Huang, Zheng Wang, Chung-Chi Tsai, Shin’ichi Satoh, and Chia-Wen Lin.
\newblock Dotscn: Group re-identification via domain-transferred single and couple representation learning.
\newblock \emph{IEEE Transactions on Circuits and Systems for Video Technology}, 31\penalty0 (7):\penalty0 2739--2750, 2020.

\bibitem[Ktena et~al.(2017)Ktena, Parisot, Ferrante, Rajchl, Lee, Glocker, and Rueckert]{ktena2017distance}
Sofia~Ira Ktena, Sarah Parisot, Enzo Ferrante, Martin Rajchl, Matthew Lee, Ben Glocker, and Daniel Rueckert.
\newblock Distance metric learning using graph convolutional networks: Application to functional brain networks.
\newblock In \emph{Medical Image Computing and Computer Assisted Intervention- MICCAI 2017: 20th International Conference, Quebec City, QC, Canada, September 11-13, 2017, Proceedings, Part I 20}, pages 469--477. Springer, 2017.

\bibitem[Lee et~al.(2019)Lee, Lee, Kim, Kosiorek, Choi, and Teh]{lee2019set}
Juho Lee, Yoonho Lee, Jungtaek Kim, Adam Kosiorek, Seungjin Choi, and Yee~Whye Teh.
\newblock Set transformer: A framework for attention-based permutation-invariant neural networks.
\newblock In \emph{ICML}, pages 3744--3753. PMLR, 2019.

\bibitem[Lin et~al.(2019)Lin, Li, Xiao, See, Zou, Xiong, Wang, and Mei]{lin2019group}
Weiyao Lin, Yuxi Li, Hao Xiao, John See, Junni Zou, Hongkai Xiong, Jingdong Wang, and Tao Mei.
\newblock Group reidentification with multigrained matching and integration.
\newblock \emph{IEEE transactions on cybernetics}, 51\penalty0 (3):\penalty0 1478--1492, 2019.

\bibitem[Lisanti et~al.(2017)Lisanti, Martinel, Del~Bimbo, and Luca~Foresti]{lisanti2017group}
Giuseppe Lisanti, Niki Martinel, Alberto Del~Bimbo, and Gian Luca~Foresti.
\newblock Group re-identification via unsupervised transfer of sparse features encoding.
\newblock In \emph{ICCV}, pages 2449--2458, 2017.

\bibitem[Lu et~al.(2022)Lu, Sheng, and Li]{lu2022tsn}
Xiaoyan Lu, Weijie Sheng, and Xinde Li.
\newblock Tsn-greid: Transformer-based siamese network for group re-identification.
\newblock In \emph{2022 17th International Conference on Control, Automation, Robotics and Vision (ICARCV)}, pages 422--427. IEEE, 2022.

\bibitem[Lu et~al.(2023)Lu, Li, Hu, Dunkin, Li, and Ge]{lu2023siamese}
Xiaoyan Lu, Xinde Li, Chuanfei Hu, Fir Dunkin, Heqing Li, and Shuzhi~Sam Ge.
\newblock Siamese transformer for group re-identification via multi-scale feature transform and joint learning.
\newblock \emph{IEEE Transactions on Instrumentation and Measurement}, 2023.

\bibitem[Murphy et~al.(2018)Murphy, Srinivasan, Rao, and Ribeiro]{murphy2018janossy}
Ryan~L Murphy, Balasubramaniam Srinivasan, Vinayak Rao, and Bruno Ribeiro.
\newblock Janossy pooling: Learning deep permutation-invariant functions for variable-size inputs.
\newblock \emph{arXiv preprint arXiv:1811.01900}, 2018.

\bibitem[Niepert et~al.(2016)Niepert, Ahmed, and Kutzkov]{niepert2016learning}
Mathias Niepert, Mohamed Ahmed, and Konstantin Kutzkov.
\newblock Learning convolutional neural networks for graphs.
\newblock In \emph{ICML}, pages 2014--2023. PMLR, 2016.

\bibitem[Qi et~al.(2018)Qi, Wang, Jia, Shen, and Zhu]{qi2018learning}
Siyuan Qi, Wenguan Wang, Baoxiong Jia, Jianbing Shen, and Song-Chun Zhu.
\newblock Learning human-object interactions by graph parsing neural networks.
\newblock In \emph{ECCV}, pages 401--417, 2018.

\bibitem[Respondek and Westwa{\'n}ska(2019)]{respondek2019counting}
Jerzy Respondek and Weronika Westwa{\'n}ska.
\newblock Counting instances of objects specified by vague locations using neural networks on example of honey bees.
\newblock \emph{Annals of Computer Science and Information Systems}, 18:\penalty0 87--90, 2019.

\bibitem[Ristani et~al.(2016)Ristani, Solera, Zou, Cucchiara, and Tomasi]{ristani2016performance}
Ergys Ristani, Francesco Solera, Roger Zou, Rita Cucchiara, and Carlo Tomasi.
\newblock Performance measures and a data set for multi-target, multi-camera tracking.
\newblock In \emph{ECCV}, pages 17--35. Springer, 2016.

\bibitem[Vaswani(2017)]{vaswani2017attention}
A Vaswani.
\newblock Attention is all you need.
\newblock \emph{NeurIPS}, 2017.

\bibitem[Wang et~al.(2022)Wang, Zhang, Lai, Xie, and Dong]{wang2022learning}
Yuan Wang, Quan Zhang, Jianhuang Lai, Xiaohua Xie, and Junhao Dong.
\newblock Learning bi-directional feature propagation with latent layout modeling for group re-identification.
\newblock In \emph{ICPR}, pages 907--913. IEEE, 2022.

\bibitem[Wang et~al.(2014)Wang, Hu, Liang, Leng, and Sun]{wang2014region}
Zheng Wang, Ruimin Hu, Chao Liang, Qingming Leng, and Kaimin Sun.
\newblock Region-based interactive ranking optimization for person re-identification.
\newblock In \emph{Advances in Multimedia Information Processing--PCM 2014: 15th Pacific-Rim Conference on Multimedia, Kuching, Malaysia, December 1-4, 2014, Proceedings 15}, pages 1--10. Springer, 2014.

\bibitem[Wang et~al.(2017)Wang, Hu, Yu, Jiang, Ma, and Satoh]{wang2017statistical}
Zheng Wang, Ruimin Hu, Yi Yu, Junjun Jiang, Jiayi Ma, and Shin'ichi Satoh.
\newblock Statistical inference of gaussian-laplace distribution for person verification.
\newblock In \emph{ACM MM}, pages 1609--1617, 2017.

\bibitem[Wu et~al.(2022)Wu, Zhu, and Gong]{wu2022learning}
Guile Wu, Xiatian Zhu, and Shaogang Gong.
\newblock Learning hybrid ranking representation for person re-identification.
\newblock \emph{PR}, 121:\penalty0 108239, 2022.

\bibitem[Wu et~al.(2021)Wu, Jain, Wright, Mirhoseini, Gonzalez, and Stoica]{wu2021representing}
Zhanghao Wu, Paras Jain, Matthew Wright, Azalia Mirhoseini, Joseph~E Gonzalez, and Ion Stoica.
\newblock Representing long-range context for graph neural networks with global attention.
\newblock \emph{NeurIPS}, 34:\penalty0 13266--13279, 2021.

\bibitem[Xiao et~al.(2018)Xiao, Lin, Sheng, Lu, Yan, Wang, Ding, Zhang, and Xiong]{xiao2018group}
Hao Xiao, Weiyao Lin, Bin Sheng, Ke Lu, Junchi Yan, Jingdong Wang, Errui Ding, Yihao Zhang, and Hongkai Xiong.
\newblock Group re-identification: Leveraging and integrating multi-grain information.
\newblock In \emph{ACM MM}, pages 192--200, 2018.

\bibitem[Xiong and Lai(2023)]{xiong2023similarity}
Jianghao Xiong and Jianhuang Lai.
\newblock Similarity metric learning for rgb-infrared group re-identification.
\newblock In \emph{CVPR}, pages 13662--13671, 2023.

\bibitem[Yan et~al.(2020)Yan, Qin, Ni, Chen, Liu, Zhu, Zheng, Yang, and Shao]{yan2020learning}
Yichao Yan, Jie Qin, Bingbing Ni, Jiaxin Chen, Li Liu, Fan Zhu, Wei-Shi Zheng, Xiaokang Yang, and Ling Shao.
\newblock Learning multi-attention context graph for group-based re-identification.
\newblock \emph{IEEE TPAMI}, 45\penalty0 (6):\penalty0 7001--7018, 2020.

\bibitem[Yu et~al.(2023)Yu, Huang, Lai, and Feng]{yu2023patch}
Lisha Yu, Sien Huang, Jianhuang Lai, and Zhanxiang Feng.
\newblock Patch-based camera-aware person-to-group learning and group similarity strategy for unsupervised group re-identification.
\newblock \emph{Neurocomputing}, 552:\penalty0 126565, 2023.

\bibitem[Zaheer et~al.(2017)Zaheer, Kottur, Ravanbakhsh, Poczos, Salakhutdinov, and Smola]{zaheer2017deep}
Manzil Zaheer, Satwik Kottur, Siamak Ravanbakhsh, Barnabas Poczos, Russ~R Salakhutdinov, and Alexander~J Smola.
\newblock Deep sets.
\newblock \emph{NeurIPS}, 30, 2017.

\bibitem[Zhang et~al.(2022{\natexlab{a}})Zhang, Dang, Lai, Feng, and Xie]{zhang2022modeling}
Quan Zhang, Kaiheng Dang, Jian-Huang Lai, Zhanxiang Feng, and Xiaohua Xie.
\newblock Modeling 3d layout for group re-identification.
\newblock In \emph{CVPR}, pages 7512--7520, 2022{\natexlab{a}}.

\bibitem[Zhang et~al.(2022{\natexlab{b}})Zhang, Lai, Feng, and Xie]{zhang2022uncertainty}
Quan Zhang, Jian-Huang Lai, Zhanxiang Feng, and Xiaohua Xie.
\newblock Uncertainty modeling with second-order transformer for group re-identification.
\newblock In \emph{Proceedings of the AAAI conference on artificial intelligence}, pages 3318--3325, 2022{\natexlab{b}}.

\bibitem[Zhang et~al.(2024)Zhang, Lai, Feng, and Xie]{zhang2024uncertainty}
Quan Zhang, Jianhuang Lai, Zhanxiang Feng, and Xiaohua Xie.
\newblock Uncertainty modeling for group re-identification.
\newblock \emph{IJCV}, pages 1--21, 2024.

\bibitem[Zhang et~al.(2021)Zhang, Wang, Cui, Zhou, Cui, Huang, and Yang]{zhang2021deep}
Tong Zhang, Yun Wang, Zhen Cui, Chuanwei Zhou, Baoliang Cui, Haikuan Huang, and Jian Yang.
\newblock Deep wasserstein graph discriminant learning for graph classification.
\newblock In \emph{AAAI}, pages 10914--10922, 2021.

\bibitem[Zheng et~al.(2009)Zheng, Gong, and Xiang]{zheng2009associating}
Wei-Shi Zheng, Shaogang Gong, and Tao Xiang.
\newblock Associating groups of people.
\newblock In \emph{BMVC}, pages 1--11, 2009.

\bibitem[Zhu et~al.(2016)Zhu, Chu, and Yu]{zhu2016consistent}
Feng Zhu, Qi Chu, and Nenghai Yu.
\newblock Consistent matching based on boosted salience channels for group re-identification.
\newblock In \emph{2016 IEEE International Conference on Image Processing (ICIP)}, pages 4279--4283. IEEE, 2016.

\bibitem[Zhu et~al.(2020)Zhu, Yang, Lin, Liu, Wang, and Zhang]{zhu2020group}
Ji Zhu, Hua Yang, Weiyao Lin, Nian Liu, Jia Wang, and Wenjun Zhang.
\newblock Group re-identification with group context graph neural networks.
\newblock \emph{IEEE Transactions on Multimedia}, 23:\penalty0 2614--2626, 2020.

\end{thebibliography}
}

 \clearpage
\setcounter{page}{1}
\maketitlesupplementary

\section{Additional Hyper-parameter Analysis}
\begin{figure}[t!]
	\centering
	\includegraphics[scale=0.35]{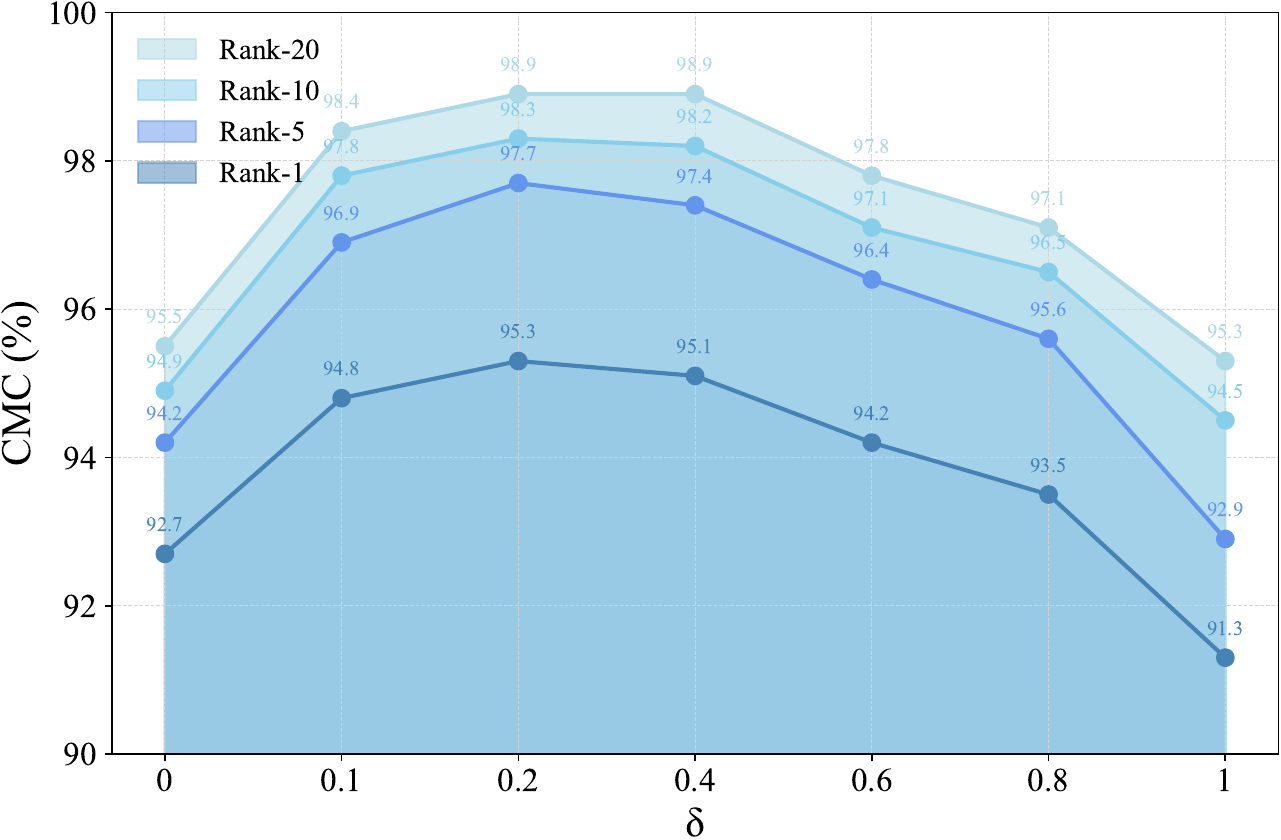}
	\caption{Analysis of the impact of reconstruction loss weight $\delta$ on the CMC at Rank-1, Rank-5, Rank-10, and Rank-20.}
	\label{fig_ap2} \vspace{-2mm}
\end{figure}

\noindent\textbf{Effect of reconstruction loss weight $\delta$}. To improve the representation of the multi-relational awareness graphs by enhancing relational information within the global-aware graph, the reconstruction loss $\mathcal{L}^{re}$ is employed to constrain the decoupling process of the global-aware graph. As shown in Fig.\ref{fig_ap2}, setting $\delta=0.0$ (disabling reconstruction loss) results in a lack of explicit guidance for preserving relational structures, leading to suboptimal graph representations and degraded performance. Increasing $\delta$ strengthens the constraints on the decoupling process, effectively enriching multi-relational information and improving performance, with the best results observed at $\delta=0.2$. However, excessively large $\delta$ introduces over-constraint, impairing the model’s adaptability to relational diversity, thereby reducing overall performance.

\begin{figure}[t!]
	\centering
	\includegraphics[scale=0.4]{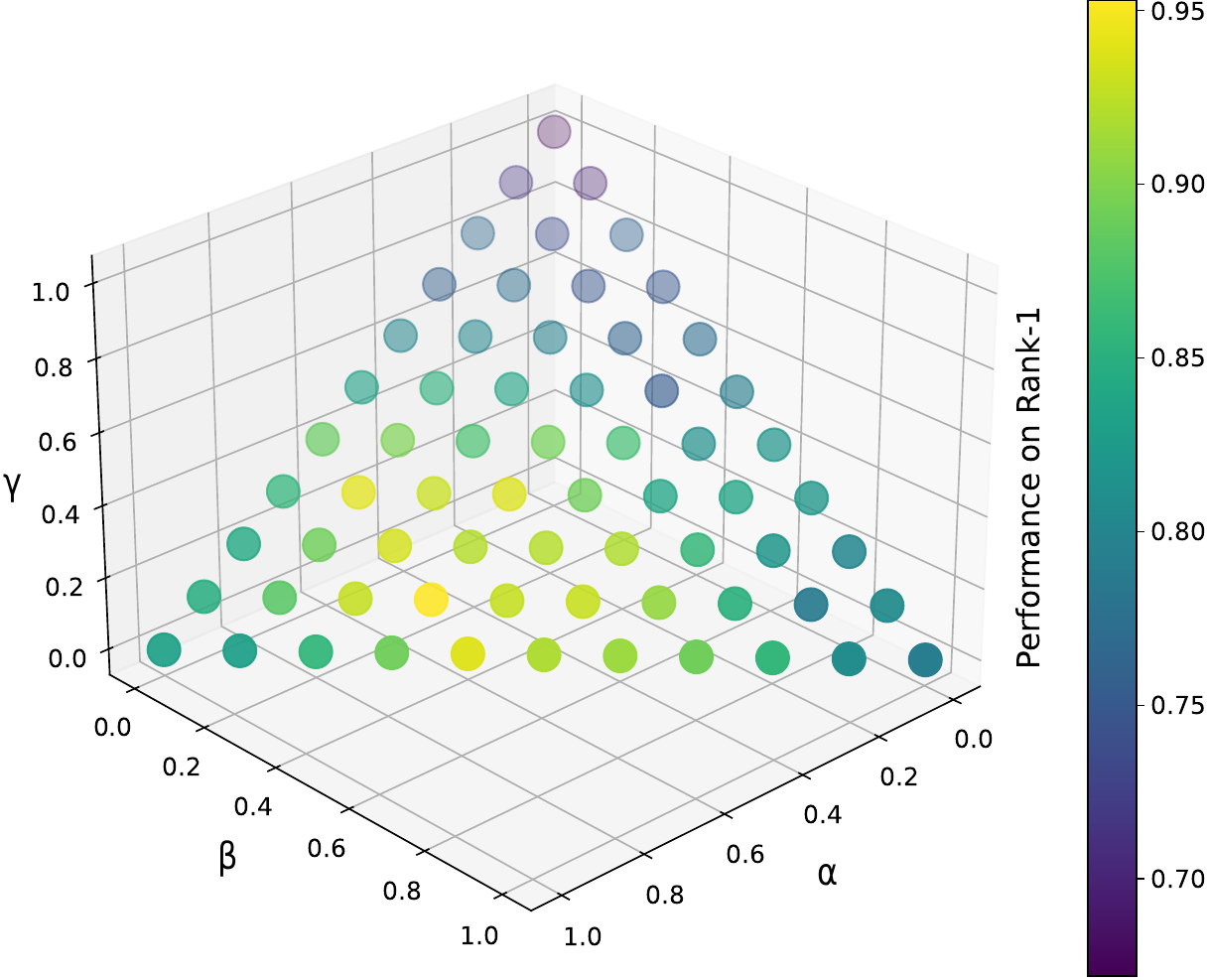}
	\caption{Analysis of the impact of multi-scale matching weighting parameters $\alpha$, $\beta$, and $\gamma$ on the Rank-1.}
	\label{fig_ap3} \vspace{-2mm}
\end{figure}

\noindent\textbf{Effect of multi-scale matching weighting parameters $\alpha$, $\beta$, and $\gamma$}. The multi-scale objective function for group matching incorporates individual-scale, local-scale, and global-scale matching, with respective contributions controlled by the weights $\alpha$, $\beta$, and $\gamma$. As shown in Fig.\ref{fig_ap3}, balancing these weights is crucial for effective matching across scales. Setting $\alpha=0.6$, $\beta=0.3$, and $\gamma=0.1$ achieves optimal performance by emphasizing individual-scale relationships while preserving local- and global-scale context. Increasing $\alpha$ further enhances individual-scale focus but diminishes the contribution of broader relational structures, reducing overall robustness. Conversely, a larger $\beta$ or $\gamma$ shifts the focus to local or global scales, which may oversimplify fine-grained associations. The chosen balance ensures tight integration of relational information across scales, maximizing group matching accuracy.

\section{Details in Occlusion Relational Modeling}
\begin{figure*}[t!]
	\centering
	\includegraphics[scale=0.13]{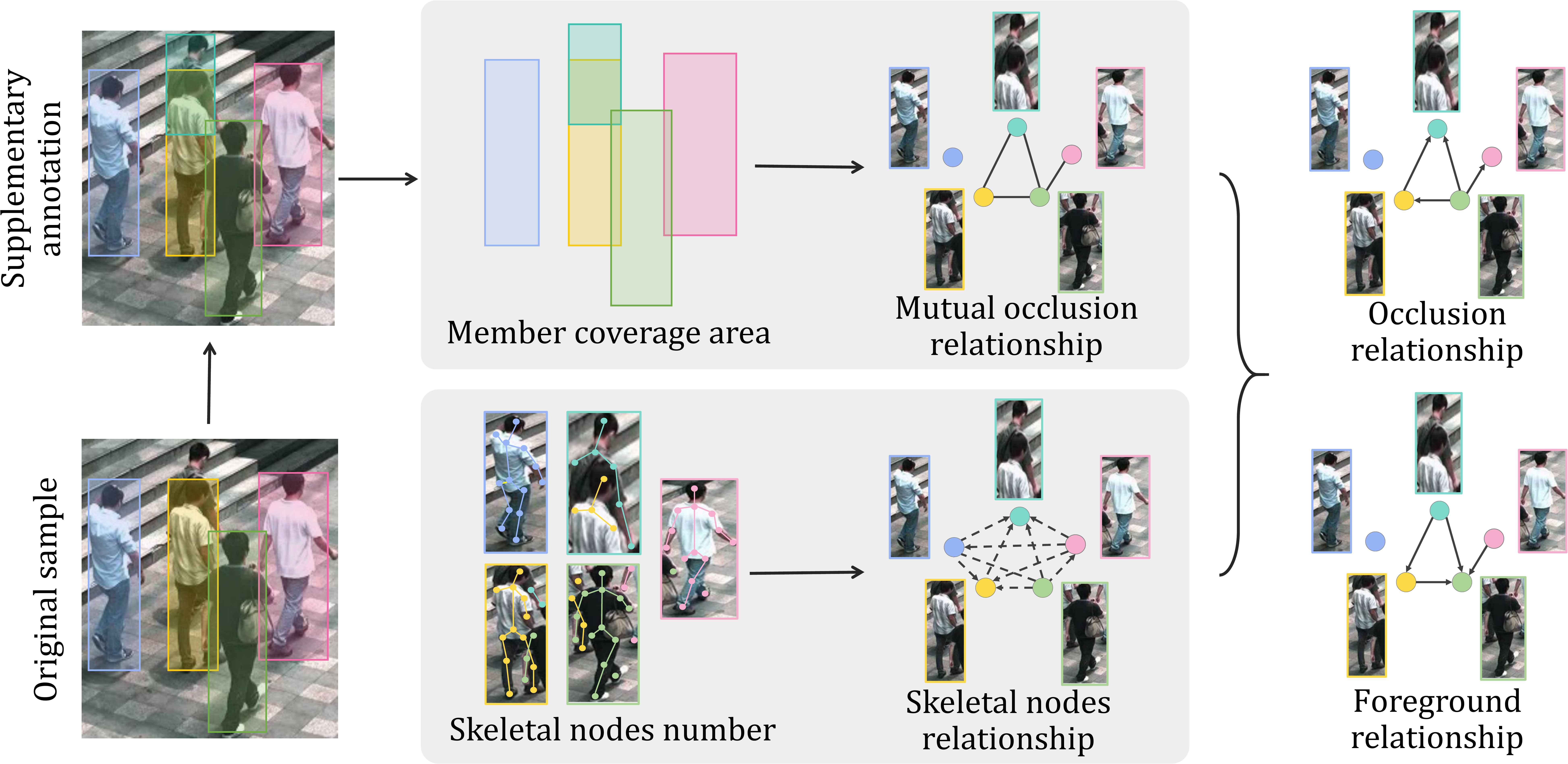}
	\caption{\textbf{The process of occlusion relational modeling}. First, we supplemented annotations for severely occluded, unlabeled members.  Next, we preliminarily determined the mutual occlusion relationships among members by analyzing their overlapping areas in the group image.  Then, we further refined the occlusion and foreground relationships based on the number of keypoints for each member.  This approach efficiently extracts occlusion and foreground information within the group with minimal resource consumption.}
	\label{fig_ap1} 
\end{figure*}

\noindent\textbf{Occlusion and foreground information extraction}. In the original datasets RoadGroup\cite{lin2019group} and CSG\cite{yan2020learning}, occlusion information among members is not directly provided. However, these datasets include the vertex pixel position information for all members $\{I_1, I_2, \dots, I_N\}$ within a given group sample $I$, represented as $\{[x_{1}^{lt}, y_{1}^{lt}, x_{1}^{rb}, y_{1}^{rb}], \dots, [x_{N}^{lt}, y_{N}^{lt}, x_{N}^{rb}, y_{N}^{rb}]\}$. As illustrated in Fig.\ref{fig_ap1}, we use the rectangular regions defined by the top-left and bottom-right vertices of each member to analyze the overlap relationships among these rectangles, preliminarily determining whether occlusion exists between specific pairs of members.

To further distinguish the occlusion and foreground relationships between members, we calculate the number of visible key points for each member. Intuitively, due to occlusion, the number of keypoints visible for members located underneath a layer is often fewer than for those located on the upper layer. Specifically, for a given pair of members $I_i$and $I_j$, we use Ultralytics\cite{ultralytics2024} pose detection algorithm to determine the skeletal nodes number $K(I_i)$and $K(I_j)$for each member. Combining this information with the overlap relationship between their respective rectangular regions, we determine whether $I_j$ is occluded by $I_i$, i.e., $I_i \in occlusion(I_j)$. The formulation is expressed as:
\begin{equation}
    occlusion(I_j)=\{I_i|R(I_i)\cap R(I_j)\ne \phi\ , K(I_i)>K(I_j) \},
\end{equation}
where $R(I_i)$ denotes the rectangular region of member $I_i$. By combining geometric overlap relationships with keypoint count information, we achieve fine-grained modeling of occlusion relationships.

\noindent\textbf{Supplementary annotations for heavily occluded members}. In the original datasets, some members with heavy occlusions were not annotated, which can result in suboptimal performance in traditional models due to incomplete member features. However, in our HMGL framework, these incomplete annotations provide valuable additional occlusion and foreground information. As demonstrated in Fig.\ref{fig_ap1}, we performed supplementary annotations for heavily occluded target members in the RoadGroup dataset, where occlusion scenarios are particularly severe. An in-depth analysis of group matching results revealed a significant gap in both accuracy and stability for samples lacking supplementary annotations compared to fully annotated ones. This underscores the critical role of supplementary occlusion relationships and member information in improving model performance. Further experiments validate the advantages of our method in handling complex re-identification tasks. As shown in Table \ref{tab_app1}, incorporating the supplemented occlusion relationships and member information significantly boosts the performance of the HMGL framework on the RoadGroup dataset.

\begin{table}[t]
    \centering  
    \caption{Effect of supplementary annotations in RoadGroup.}
    \small 
    \setlength{\tabcolsep}{4pt} 
    \begin{tabular}{c|c|c|c|c|c}
    \hline
    \multirow{2}{*}{Dataset} & \multicolumn{5}{c}{HMGL} \\ \cline{2-6} 
                          & Rank-1 & Rank-5 & Rank-10 & Rank-20 & mAP \\ \hline \hline
    Original                     & 92.2   & 97.5   & 98.8    & 99.2    & 94.1 \\ \hline
    Supplemented                     & \textbf{93.9} & \textbf{98.4} & \textbf{99.0} & \textbf{99.2} & \textbf{95.4} \\
    \hline
    \end{tabular}
    \label{tab_app1} \vspace{-3mm}
\end{table}

\section{Additional Qualitative Results}
To further analyze the performance of our method, we conducted additional qualitative experiments, extending the visualization of group queries and including retrieval results from Rank-1 to Rank-10, as shown in Fig.\ref{fig_results}. The experimental results still reveal that in the graph construction test, relying solely on explicit or implicit graph methods exhibited significant instability across different samples. In the group matching test, global matching methods showed limited effectiveness in handling high-difficulty samples, while node matching methods demonstrated some improvement but still fell short of achieving optimal performance.

\begin{figure*}[t!]
	\centering
	\includegraphics[scale=0.21]{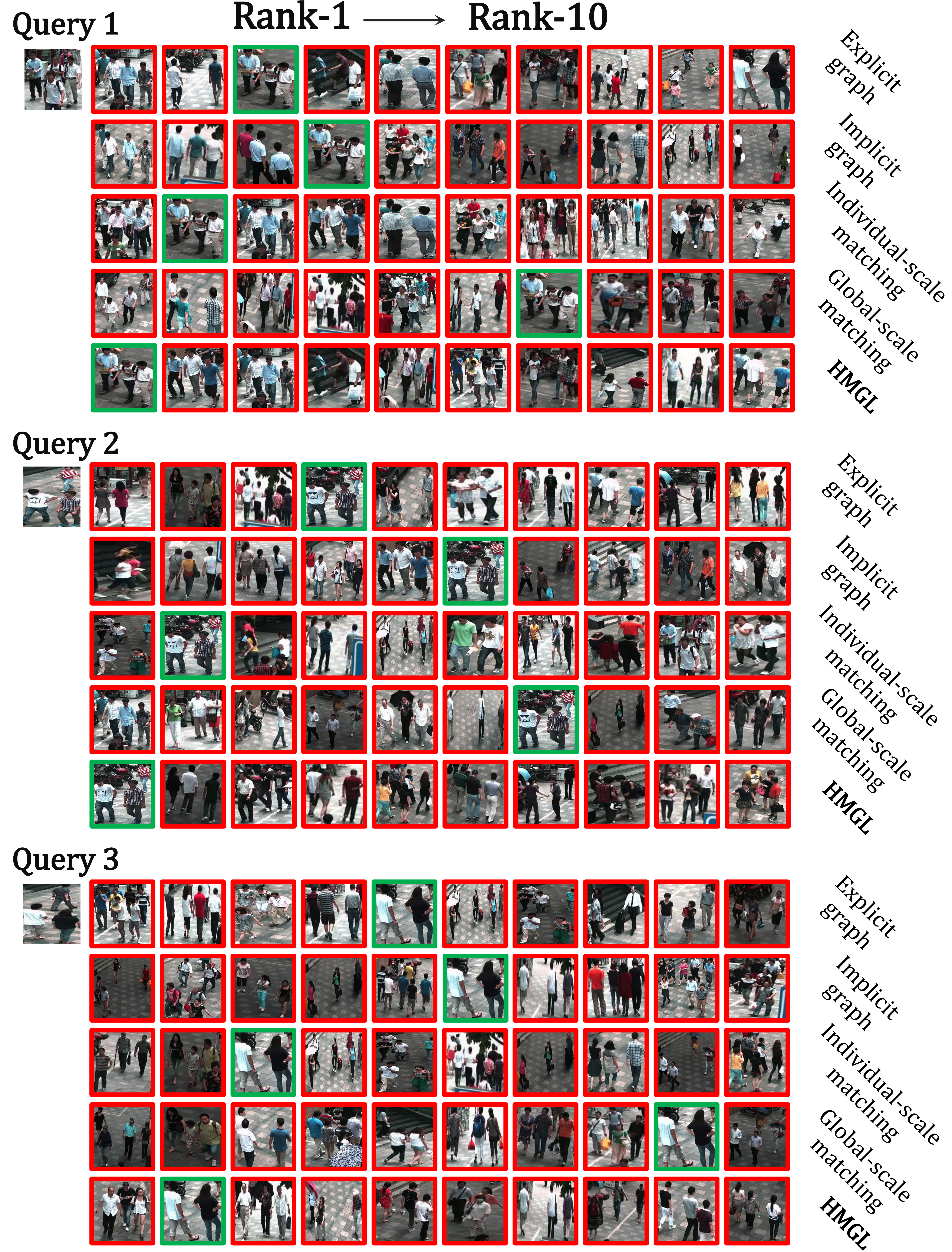}
	\label{fig_results} 
\end{figure*}

\begin{figure*}[t!]
	\centering
	\includegraphics[scale=0.21]{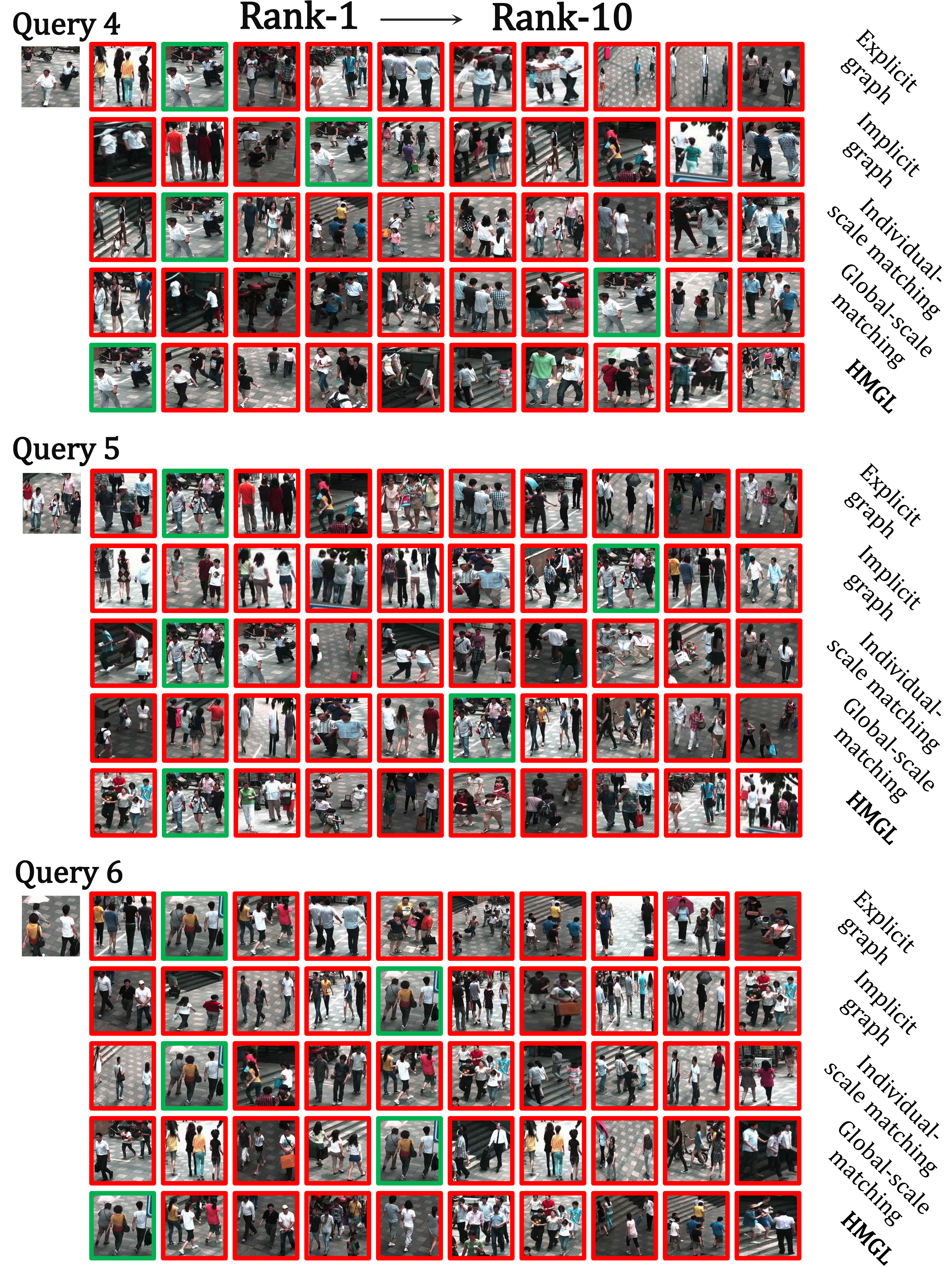}
	\caption{\textbf{Comparative analysis and qualitative experiments for G-ReID}.Query examples (1-6) show Rank-1 to Rank-5 retrieval results, with green-bordered images indicating correct matches and red-bordered ones indicating incorrect matches.  The extended analysis includes Rank-1 to Rank-10 retrieval, highlighting the instability of general graph construction methods and the limitations of global matching for complex samples.}
	\label{fig_results} 
\end{figure*}
 

\end{document}